\pdfoutput=1

\documentclass[11pt]{article}

\usepackage[]{EMNLP2022}

\usepackage{times}
\usepackage{latexsym}

\usepackage[T1]{fontenc}

\usepackage[utf8]{inputenc}

\usepackage{microtype}

\usepackage{inconsolata}

\usepackage{booktabs}
\usepackage{graphicx}
\usepackage{multirow}
\usepackage{todonotes}

\newcommand{\tommaso}[1]{\textcolor{black}{#1}}
\newcommand{\gosse}[1]{\textcolor{black}{#1}}
\newcommand{\fnframe}[1]{\textsc{#1}}
\newcommand{\sara}[1]{\textcolor{black}{#1}}
\newcommand{\rebuttal}[1]{#1}

\title{Dead or Murdered? Predicting Responsibility \\Perception in Femicide News Reports}

\author{Gosse Minnema$^a$, Sara Gemelli$^b$, Chiara Zanchi$^b$,\\\textbf{Tommaso Caselli$^a$,} \textbf{Malvina Nissim$^a$} \AND \textnormal{$^a$University of Groningen, The Netherlands}\\$^b$University of Pavia, Italy\\\texttt{g.f.minnema@rug.nl}}

\begin{document}
\maketitle

\begin{abstract}
Different linguistic expressions can conceptualize the same event from different viewpoints by emphasizing certain participants over others. Here, we investigate a case where this has social consequences: how do linguistic expressions of gender-based violence (GBV) influence who we perceive as responsible? We build on previous psycholinguistic research in this area and conduct a large-scale perception survey of GBV descriptions automatically extracted from a corpus of Italian newspapers. We then train regression models that predict the salience of GBV participants with respect to different dimensions of perceived responsibility. Our best model (fine-tuned BERT) shows solid overall performance, with large differences between dimensions and participants: salient \textit{focus} is more predictable than salient \textit{blame}, and perpetrators' salience is more predictable than victims' salience. Experiments with ridge regression models using different representations show that features based on linguistic theory perform similarly to word-based features. Overall, we show that different linguistic choices do trigger different perceptions of responsibility, and that such perceptions can be modelled automatically. This work can be a core instrument to raise awareness of the consequences of different perspectivizations in the general public and in news producers alike. 
\end{abstract}

\section{Introduction and background}
\label{sec:introduction}

The same event can be described in many different ways, according to who reports on it, and the choices they make. They can opt for some words rather than others, for example, or they can use a passive rather than an active construction, or more widely, they can -- consciously or not -- provide the reader with a specific perspective over what happened.

\begin{figure}[t]
    \centering
    \includegraphics[width=\columnwidth]{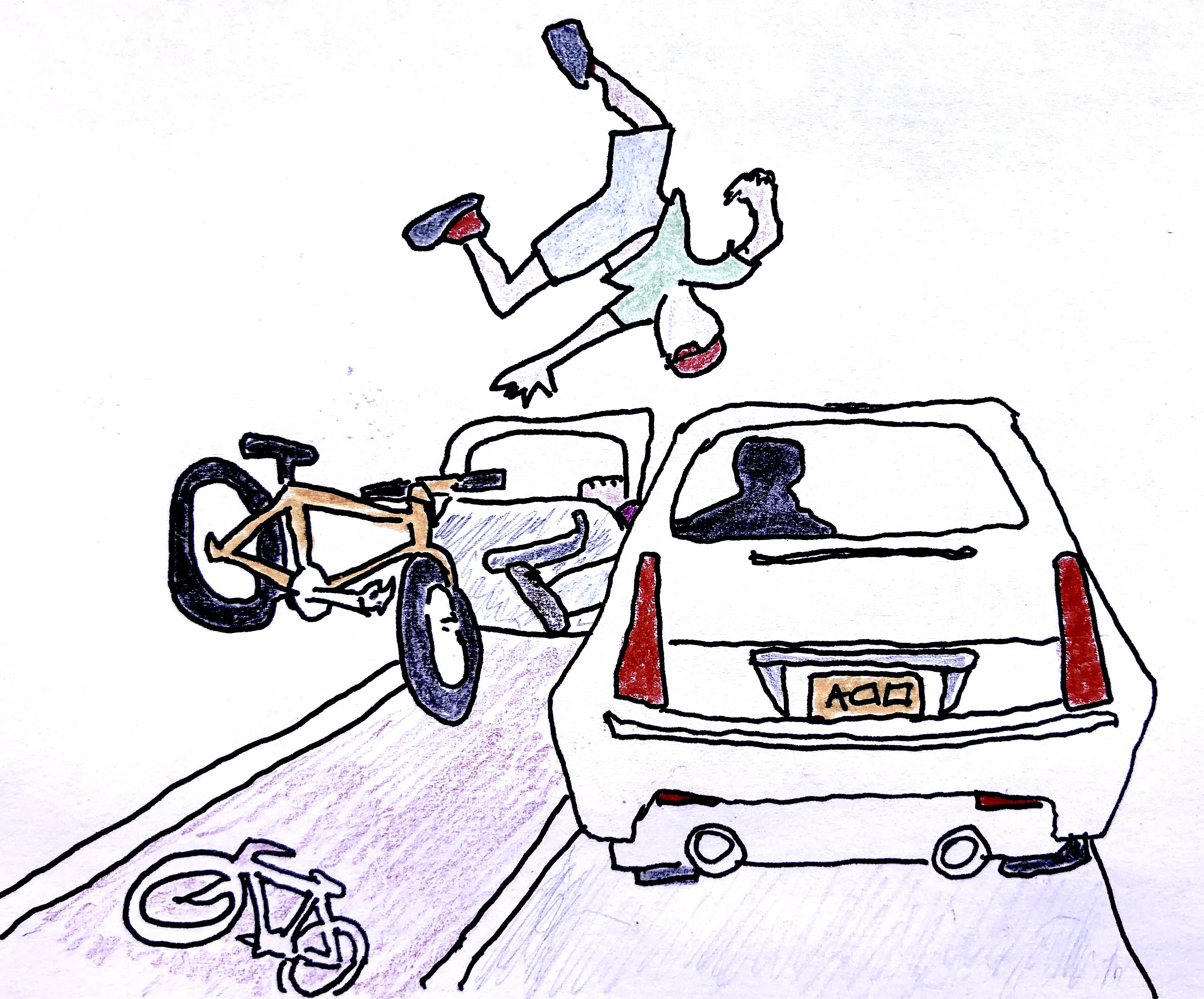}
    \caption{
    ``Cyclist slams into car door''\\
    Figure~1: ``Car driver opens door and hits cyclist''\\
    Figure~1: ``Cyclist injured in road accident on 5th Street''\\
    Figure~1: ``Collision between bike and car''\\
    \textit{We use alternative captions to illustrate how the same event can be described from alternative perspectives, which can evoke different perceptions in the attribution of responsibility to the actors involved.}
    \label{fig:perspective}}
\end{figure}

Such choices do not just pertain to the realm of stylistic subtleties; rather, they can have substantial consequences on how we think of -- or \textit{frame} -- events. Indeed, it is known that the way a piece of news is written, especially in terms of perspective-taking, heavily influences the way readers perceive \textit{attribution of responsibility} in the events described \citep{iyengar1994anyone}.
Figure~\ref{fig:perspective}\footnote{Drawing inspired by the illustration in \url{https://www.outsideonline.com/culture/opinion/look-you-open-your-car-door/} (accessed 2022-09-22).} illustrates how the same event can be reported on from different viewpoints, in ways that do affect the perception of the participants' responsibilities. 
We are interested in unpacking \textit{responsibility attribution} using NLP tools in the context of a socially relevant phenomenon, namely gender-based violence (GBV).

Violence against women is worryingly common and therefore often reported in the news. A report by the European parliament \cite{ep-femicides-2021} 
details an estimate of 87,000 women intentionally killed in 2017. While Italy is listed in this report as one of the European countries with the lowest number of femicides, they are still too frequent and have been constant in the last 25 years (0.6 per 100,000 women in 1982 and 0.4 per 100,000 in 2017). Most discouragingly, a report from November 2018 by 
two Italian research institutes
points out that the stereotype of a shared responsibility between the violence victim and its perpetrator is still widespread among young generations: ``56.8\% of boys and 38.8\% of girls believe that the female is at least partly responsible for the violence she has suffered" \cite{la-iard-adolescenti-2018}.

Working on Italian news, \citet{pinelli2021gender} observe that in descriptions of femicides, the use of syntactic constructions with varying levels of transitivity -- from transitive active constructions on one side of the spectrum, via passives and anticausatives to nominalization constructions on the other side -- corresponds to various degrees of responsibility attributed to the (male) perpetrator. For example, while ``\textit{he killed her}'' (active/transitive) makes the involvement of an active agent fully explicit, with ``\textit{she was killed (by him)}'' (passive) the event is accessed via the patient shifting attention away from the agent, and expressions such as ``\textit{the murder}'' or even ``\textit{the event}'' (nominal construction) moves both participants to the background. 
\sara{In a related contribution, \citet{meluzzi2021responsibility} investigate the impact of argument structure constructions on responsibility attributions by means of a survey on artificially-constructed GBV reports in Italian. Their results further confirm the findings of \citet{pinelli2021gender} on the effects of readers' perception on the agentivity and responsibility of the perpetrators and the victims.} The outcomes of both studies is in line with previous work in psycholinguistics showing that in events involving violence (at any
level), the linguistic backgrounding of agents hinders
their responsibility and promotes victim blaming~\cite{huttenlocher1968,henley1995syntax,bohner2002,gray-wegner2009,hart2020objectification,zhou-etal-2021-assessing}.


Based on such framing choices, how will the general reader perceive the described event? Can we model such perceptions automatically? In this paper we aim to answer these questions, still focusing on descriptions of femicides in Italian news, and exploiting \textit{frame semantics} \cite{fillmore2006} 
as a theoretical and practical tool, as well as most recent NLP approaches.



Using specific pre-selected semantic frames, automatically extracted using a state-of-the-art semantic parser~\cite{xia-etal-2021-lome}, we identify descriptions of GBV events from Italian newspapers. On these descriptions we collect human judgements through a large-scale survey where we ask participants to read the texts and ascribe a degree of \textit{perceived responsibility} to the perpetrator, the victim, or to some more abstract concept (e.g. ``jealousy", ``rage"). More details are provided in \S\ref{sec:questionnaire}.

Next, we model perception of responsibility automatically by developing a battery of regression models (both from scratch as well as atop pre-trained transformer models) exploiting a variety of linguistic cues which range from surface to frame-based features. The training objective of such models is the prediction of the human perception scores.
We achieve a strong correlation with a transformer-based model. 
The fine-grained character both of the survey and the result analysis that we conducted also allows us to observe differences in prediction complexity for the various aspects that we consider. Modeling and evaluation are discussed in \S\ref{sec:prediction}. 

The results we obtain show that \textbf{different linguistic choices do indeed trigger different perceptions of responsibility, and that such perceptions can be modelled automatically.} 
This finding not only confirms previous research which was conducted (manually) on a much smaller scale, but also opens up the possibility \tommaso{to conduct large-scale analyses of texts exposing to both producers and consumers of texts which perspectivization strategies are at play and their effects.\footnote{Our data and code are available at \url{gitlab.com/sociofillmore/perceived-perspective-prediction}.}}



\section{Femicide perception dataset}
\label{sec:questionnaire}

We designed an online questionnaire study in which participants \gosse{were presented with} sentences extracted from \gosse{the RAI Femicides Corpus \cite{belluati2021femminicidio},
a collection of 2,734 news articles covering 937 confirmed femicide cases perpetrated in Italy
in 2015-2017, and asked to rate the level of agentivity and responsibility expressed} in each sentence.
The results of the questionnaire demonstrate a clear effect of semantic frames and syntactic constructions on the perception of descriptions of femicides. 

\subsection{Question formulation}

The level of responsibility ascribed to event participants can be expressed in multiple ways triggering different perceptions in the readers.
Since responsibility is a complex concept, we break it down into three dimensions in order to make it (i) more understandable for our participants, and (ii) to get a more nuanced picture of readers' perceptions. The three dimensions are:

\begin{enumerate}
  \setlength{\itemsep}{1pt}
  \setlength{\parskip}{0.5pt}
  \setlength{\parsep}{0pt}
    \item FOCUS: does the sentence \textit{focus} on the agent or on something else?
    \item CAUSE: does the sentence describe the event as being \textit{caused} primarily by a human or by something else? 
    \item BLAME: does the sentence attribute \textit{blame} to the agent or to something else?
\end{enumerate}

\begin{table}[h]
\resizebox{\columnwidth}{!}{%
\begin{tabular}{@{}lccc@{}}
\toprule
\multirow{2}{*}{\textbf{Example}} & \textbf{\small{FOCUS}} & \textbf{\small{CAUSE}} & \textbf{\small{BLAME}} \\
 & \multicolumn{3}{c}{\textit{\textbf{\small ascribed to the murderer}}} \\ \midrule
Her fiancé brutally murdered her & $+$ & $+$ & $+$ \\
Blinded by jealousy, he killed her & $+$ & $+$ & $\pm$ \\
Her husband's jealousy killed her & $+$ & $-$ & $\pm$ \\
Her blind love for him became fatal & $\pm$ & $-$ & $-$ \\
A tragic incident occurred in Rome & $-$ & $-$ & $-$ \\ \bottomrule
\end{tabular}%
}
\caption{Hypothesized perceptual ratings relative to the murderer (examples are artificial)}
\label{tab:exa-ratings}
\end{table}

Table~\ref{tab:exa-ratings} shows hypothesized ratings on these dimensions for a number of artificial examples, demonstrating that the three dimensions are closely related, but do not always match: for example, the first and second sentences both focus on the role of the murderer and describe his actions as the cause, but the second sentence arguably attributes less blame to the murderer by describing him as `blinded' by jealousy, implying that he does not bear full responsibility to his actions. Note that the ratings presented in the table merely represent a hypothesis about how the sentences are likely to be perceived; perception is inherently subjective and these examples should not be taken as a `gold standard' of any kind.

To put the amount of responsibility attributed to the murderer in perspective, we also asked readers about the perceived level of focus, causation, and blame placed on the victim, an object (e.g. a weapon), a concept or emotion (e.g. jealousy), or on nothing at all. For a given sentence, participants were asked to give ratings on a 5-point Likert scale to each of these categories. Participants also had the option to indicate that the sentence was irrelevant and skip answering it. The full set of questions is given in Table~\ref{tab:questions}. Note that, taking into account preliminary results from a pilot study, the categories have been adapted slightly to each individual question: for example, we omitted the `none' category for the focus dimension (since there always has to be focus on something), and in the `cause' dimension we made the descriptions of each category slightly more elaborate.

\begin{table*}[h!]
\resizebox{\textwidth}{!}{%
\begin{tabular}{@{}lllllll@{}}
\toprule
\textbf{Dimension} & \textbf{Question} & \textbf{Murderer} & \textbf{Victim} & \textbf{Object} & \textbf{Concept} & \textbf{None} \\ \midrule
\textbf{FOCUS} & \textit{\begin{tabular}[c]{@{}l@{}}La frase concentra \\l'attenzione principalmente...\end{tabular}} & \textit{sull'assassino} & \textit{sulla vittima} & \textit{su un oggetto} & \textit{\begin{tabular}[c]{@{}l@{}}su un concetto astratto \\ o un'emozione\end{tabular}} & - \\
 & \begin{tabular}[c]{@{}l@{}}`The sentence puts most \\ attention ...'\end{tabular} & `on the assassin' & `on the victim' & `on an object' & \begin{tabular}[c]{@{}l@{}}`on an abstract concept \\ or emotion'\end{tabular} & - \\
 &  &  &  &  &  &  \\
\textbf{CAUSE} & \textit{\begin{tabular}[c]{@{}l@{}}La morte della donna è \\ descritta come ...\end{tabular}} & \textit{\begin{tabular}[c]{@{}l@{}}causata da un essere \\ umano\end{tabular}} & - & \textit{\begin{tabular}[c]{@{}l@{}}causata da un oggetto\\ (es. una pistola)\end{tabular}} & \textit{\begin{tabular}[c]{@{}l@{}}causata da un'emozione\\ (es. gelosia)\end{tabular}} & \textit{\begin{tabular}[c]{@{}l@{}}spontanea, priva di un \\ agente scatenante\end{tabular}} \\
 & \begin{tabular}[c]{@{}l@{}}`The murder of the woman \\ is described as ... '\end{tabular} & \begin{tabular}[c]{@{}l@{}}`caused by a human \\ being'\end{tabular} & - & \begin{tabular}[c]{@{}l@{}}`caused by an object \\ (e.g. a gun)'\end{tabular} & \begin{tabular}[c]{@{}l@{}}`caused by an emotion\\ (e.g. jealousy)'\end{tabular} & \begin{tabular}[c]{@{}l@{}}`spontaneous, without\\ a triggering agent'\end{tabular} \\
 &  &  &  &  &  &  \\
\textbf{BLAME} & \textit{La frase accusa...} & \textit{l'assassino} & \textit{la vittima} & \textit{un oggetto} & \textit{\begin{tabular}[c]{@{}l@{}}un concetto astratto\\ o un'emozione\end{tabular}} & \textit{nessuno} \\
\textbf{} & `The sentence accuses ...' & `the murderer' & `the victim' & `an object' & \begin{tabular}[c]{@{}l@{}}`an abstract concept \\ or an emotion'\end{tabular} & `no one' \\ \bottomrule
\end{tabular}%
}
\caption{Question dimensions and attributes}
\label{tab:questions}
\end{table*} 

\subsection{Sentence selection}
Relevant sentences were extracted from the corpus following a two-step process: First, occurrences of semantic frames were automatically extracted using the LOME parser \cite{xia-etal-2021-lome}. This information was combined with an automatic dependency parse using SpaCy \cite{spacy} to classify syntactic constructions. For example, \textit{he murdered her} would be classified as ``\fnframe{Killing}:active'' (\fnframe{Killing} frame, expressed with active syntax), \textit{she died} as ``\fnframe{Death}:intransitive'', and \textit{the tragedy} as ``\fnframe{Catastrophe}:nonverbal''.\footnote{In this context, ``nonverbal'' means `without a verb'; in this example, \textit{tragedy} is an event expressed by a noun.} In a second step, we selected \textit{typical frames} \cite{vossen-etal-2020-large} that encode possible ways of expressing the murder event with various degrees of emphasis on the various participants, and randomly sampled sentences containing at least one of these frames. \rebuttal{Typical frames were selected by manually annotating the example sentences from \citet{pinelli2021gender} with FrameNet frames, and selecting the frames evoked by words that refer to (or imply) the event of the death of the victim (``he \textit{killed} her`` she \textit{died}``, ``she was found \textit{dead}``, ``a tragic \textit{incident}``). This yielded the set of frames $\{$\fnframe{Killing}, \fnframe{Death}, \fnframe{Dead\_or\_alive}, \fnframe{Event}, \fnframe{Catastrophe} $\}$, all of which can be used to describe exactly the same event but with different levels of dynamism (being dead vs. dying), agentivity (killing vs. dying), and generality (someone dying vs. something happening). We excluded frames that refer to events that are related to but distinct from the murder itself, such as \fnframe{Cause\_harm} and \fnframe{Use\_firearm} (``he \textit{stabbed} her'', ``he \textit{fired} his gun'' -- these may refer to the cause of death, but do not include the death itself), or
\fnframe{Offenses} (``he was charged with \textit{murder}'' -- this refers to the crime as a judicial concept, not as a real-world event).} We then sampled sentences from our corpus in such a way that we created a corpus with an equal number of examples of each frame-construction pair, and equal numbers of headlines and body-text sentences.

\subsection{Practical implementation}

Given the considerable cognitive load of analyzing (sometimes complex) sentences as well as the emotional load of reading text about a heavy and distressing topic, participants were asked to provide ratings on only one dimension, for a set of 50 sentences. Furthermore, attempting to find a balance between the \textit{depth} (number of annotations per sentence) and \textit{breath} (total number of annotations) of our annotations, we decided to set a target of 10 participants for each sentence and each dimension, meaning that 30 participants are needed to fully annotate each block of 50 sentences. \\
\indent In order to distribute participants evenly across sentence sets and dimensions, without knowing the response rate in advance, we created 60 groups (20 sets of 50 sentences [= 1,000 in total] $\times$ three dimensions) and assigned participants to groups on a rolling basis: one group was open at a time, and once the required number of participants was reached, it was automatically closed and the next group was opened.
Once a group was full, we manually inspected the responses for completeness and quality. Due to the subjective nature of the task, there are no `wrong' responses per se, but we considered responses to be of low quality if they met at least one of the following three criteria: (i) implausibly fast completion of the questionnaire,\footnote{We considered responses `too fast' if they took less than 6 minutes (for 50 sentences, i.e. 7 sec./sentence, not including time spent reading instructions).} (ii) suspicious patterns of marking sentences as irrelevant and skipping them (e.g. skipping many sentences in a row), or (iii) suspicious response patterns (e.g. always giving the same ratings to each sentence). 

The link to the survey platform\footnote{We used Qualtrics (\url{https://www.qualtrics.com/}) to present stimuli and collect responses, alongside an in-house system for managing participants and payments.} was distributed amongst university students enrolled in bachelor's and master's degrees in different programs at several universities in Italy. Responses were collected anonymously, but participants were asked to state their gender, age, and profession. 

\subsection{Results}
\label{sub:questionnaire-results}

\begin{table}[]
\centering
\resizebox{.48\textwidth}{!}{%
\begin{tabular}{@{}llrrrrrrr@{}}
\toprule
\multicolumn{2}{r}{\textbf{participant}} & \multicolumn{2}{c}{\textit{all}} & \multicolumn{2}{c}{\textit{female}} & \multicolumn{2}{c}{\textit{male}} \\
\multicolumn{2}{r}{\textbf{scores}} & \textit{mean} & \textit{std} & \textit{mean} & \textit{std} & \textit{mean} & \textit{std} \\ \midrule
\textit{blame} & \textit{murderer} & \cellcolor[HTML]{B1E0C9}2.35 & \cellcolor[HTML]{FFE69F}1.89 & \cellcolor[HTML]{BAE3CF}2.07 & \cellcolor[HTML]{FFE7A4}1.80 & \cellcolor[HTML]{A3DABF}2.75 & \cellcolor[HTML]{FFE499}2.01 \\
\textit{} & \textit{victim} & \cellcolor[HTML]{EFF9F4}0.49 & \cellcolor[HTML]{FFF3D1}0.92 & \cellcolor[HTML]{F1FAF5}0.44 & \cellcolor[HTML]{FFF3D1}0.92 & \cellcolor[HTML]{EDF8F3}0.55 & \cellcolor[HTML]{FFF3D0}0.92 \\
\textit{} & \textit{object} & \cellcolor[HTML]{F0F9F5}0.46 & \cellcolor[HTML]{FFF2CC}1.01 & \cellcolor[HTML]{F1FAF5}0.44 & \cellcolor[HTML]{FFF2CC}1.02 & \cellcolor[HTML]{EFF9F4}0.50 & \cellcolor[HTML]{FFF2CD}0.99 \\
\textit{} & \textit{concept} & \cellcolor[HTML]{E4F4EC}0.82 & \cellcolor[HTML]{FFEEBD}1.30 & \cellcolor[HTML]{E4F4EC}0.83 & \cellcolor[HTML]{FFEDBB}1.33 & \cellcolor[HTML]{E5F5ED}0.79 & \cellcolor[HTML]{FFEEC0}1.25 \\
\textit{} & \textit{no-one} & \cellcolor[HTML]{D2EDE0}1.36 & \cellcolor[HTML]{FFE8A7}1.74 & \cellcolor[HTML]{CEEBDD}1.49 & \cellcolor[HTML]{FFE7A6}1.76 & \cellcolor[HTML]{D8EFE4}1.19 & \cellcolor[HTML]{FFE8A8}1.71 \\ \midrule
\textit{cause} & \textit{human} & \cellcolor[HTML]{8AD0AD}3.51 & \cellcolor[HTML]{FFE9AA}1.68 & \cellcolor[HTML]{89CFAD}3.54 & \cellcolor[HTML]{FFE9AA}1.67 & \cellcolor[HTML]{8BD0AE}3.48 & \cellcolor[HTML]{FFE8AA}1.69 \\
\textit{} & \textit{object} & \cellcolor[HTML]{D2EDE0}1.37 & \cellcolor[HTML]{FFE6A1}1.85 & \cellcolor[HTML]{D2EDE0}1.36 & \cellcolor[HTML]{FFE6A2}1.84 & \cellcolor[HTML]{D0ECDF}1.40 & \cellcolor[HTML]{FFE59E}1.91 \\
\textit{} & \textit{concept} & \cellcolor[HTML]{E3F4EB}0.86 & \cellcolor[HTML]{FFEDBC}1.32 & \cellcolor[HTML]{E2F4EB}0.88 & \cellcolor[HTML]{FFEEBD}1.31 & \cellcolor[HTML]{E6F5EE}0.76 & \cellcolor[HTML]{FFEDBB}1.34 \\
\textit{} & \textit{no-one} & \cellcolor[HTML]{CAEADA}1.59 & \cellcolor[HTML]{FFEAAF}1.59 & \cellcolor[HTML]{CAEADA}1.58 & \cellcolor[HTML]{FFEAAE}1.59 & \cellcolor[HTML]{C9EADA}1.61 & \cellcolor[HTML]{FFEAAF}1.58 \\ \midrule
\textit{focus} & \textit{murderer} & \cellcolor[HTML]{B4E1CB}2.26 & \cellcolor[HTML]{FFE59D}1.94 & \cellcolor[HTML]{B5E1CB}2.23 & \cellcolor[HTML]{FFE59E}1.91 & \cellcolor[HTML]{B2E0CA}2.30 & \cellcolor[HTML]{FFE59B}1.97 \\
\textit{} & \textit{victim} & \cellcolor[HTML]{A0D9BD}2.85 & \cellcolor[HTML]{FFEAAE}1.60 & \cellcolor[HTML]{A6DBC1}2.68 & \cellcolor[HTML]{FFEAAE}1.59 & \cellcolor[HTML]{98D6B8}3.07 & \cellcolor[HTML]{FFEAAE}1.61 \\
\textit{} & \textit{object} & \cellcolor[HTML]{D2EDE0}1.35 & \cellcolor[HTML]{FFE9AB}1.65 & \cellcolor[HTML]{D3EDE0}1.33 & \cellcolor[HTML]{FFE9AC}1.65 & \cellcolor[HTML]{D1EDDF}1.39 & \cellcolor[HTML]{FFE9AB}1.65 \\
\textit{} & \textit{concept} & \cellcolor[HTML]{C8E9D9}1.65 & \cellcolor[HTML]{FFE9AB}1.65 & \cellcolor[HTML]{CBEADB}1.56 & \cellcolor[HTML]{FFE8A9}1.69 & \cellcolor[HTML]{C5E8D6}1.76 & \cellcolor[HTML]{FFEAAE}1.59 \\ \bottomrule
\end{tabular}%
}
\caption{Summary of perception scores per question and attribute}\label{tab:questionnaire-summary}
\end{table}

Our final dataset covers 400 sentences with ratings from 240 participants in total (153 identifying as female, 86 as male, 1 as non-binary; mean age 23.4). In Table~\ref{tab:questionnaire-summary}, a summary of the perception scores aggregated across sentences is given. We give both the mean score (in green, on a scale from 0-5), averaged over all participants and all sentences, and the standard deviation of averaged scores across sentences. Overall, the attributes corresponding to the perpetrator tend to have higher average scores but also more variance than the other attributes (except \textit{focus/victim}, which has a higher average but lower variance). More details about the distribution of scores per question and attribute are given in the Appendix. \rebuttal{Due to the inherently subjective nature of the task, and in line with previous studies on perceptual norms (e.g., \citealt{brysbaert2014}), we did not calculate inter-annotator agreement scores.}

Table~\ref{tab:questionnaire-results} (reproduced from \citealt{minnema-etal-2022-sociofillmore}) shows average scores for the \textit{focus} question, split by typical frame and construction. This shows significant effects: sentences containing the \fnframe{Killing} frame tend to put higher focus on the murderer, and substantially more so when using an active construction. Meanwhile, the use of the \fnframe{Catastrophe}, \fnframe{Dead\_or\_alive}, and \fnframe{Death} frames, as well as the \fnframe{Killing} frame used in an active or passive construction increases the focus on the victim. On the other hand, there were no significant differences in focus scores for the object, and significant but smaller differences in focus on a concept or emotion. In each of these cases, the findings correspond to what we expected based on linguistic theory: if an event participant is lexically encoded in the predicate and syntactically required to be expressed, it is more likely that this participant will be perceived as being under focus. More focus on the murderer and the victim was also expected, both based on the content of the sentences, and on the fact that several frames (e.g. \fnframe{Killing}) lexically encode the presence of a victim and/or a killer, but not necessarily that of an inanimate concept or emotion (possibly except \fnframe{Catastrophe}). 

\begin{table}[]
\resizebox{\columnwidth}{!}{%
\begin{tabular}{@{}llrrrrr@{}}
\toprule
\multicolumn{2}{l}{\textbf{frame/construction}} & \multicolumn{1}{l}{\textbf{murderer**}} & \multicolumn{1}{l}{\textbf{victim**}} & \multicolumn{1}{l}{\textbf{object}} & \multicolumn{1}{l}{\textbf{\begin{tabular}[c]{@{}l@{}}concept /\\ emotion*\end{tabular}}} \\ \midrule
\multicolumn{2}{l}{\fnframe{Catastrophe}} &  &  &  &  \\
\textbf{} & nonverbal & \cellcolor[HTML]{FFFFFF}1.319 & \cellcolor[HTML]{A4DBC0}2.713 & \cellcolor[HTML]{FFFFFF}0.760 & \cellcolor[HTML]{FFFFFF}2.190 \\
\multicolumn{2}{l}{\fnframe{Dead\_or\_alive}} &  &  &  &  \\
\textbf{} & nonverbal & \cellcolor[HTML]{FFFFFF}1.195 & \cellcolor[HTML]{8ED1B0}3.387 & \cellcolor[HTML]{FFFFFF}1.386 & \cellcolor[HTML]{FFFFFF}1.993 \\
\textit{} & intransitive & \cellcolor[HTML]{FFFFFF}1.983 & \cellcolor[HTML]{89D0AD}3.529 & \cellcolor[HTML]{FFFFFF}1.566 & \cellcolor[HTML]{FFFFFF}1.539 \\
\multicolumn{2}{l}{\fnframe{Death}} & \multicolumn{1}{l}{} & \multicolumn{1}{l}{} & \multicolumn{1}{l}{} & \multicolumn{1}{l}{} \\
\textit{} & nonverbal & \cellcolor[HTML]{FFFFFF}0.967 & \cellcolor[HTML]{92D3B4}3.247 & \cellcolor[HTML]{FFFFFF}1.507 & \cellcolor[HTML]{FFFFFF}1.914 \\
\textit{} & intransitive & \cellcolor[HTML]{FFFFFF}1.867 & \cellcolor[HTML]{7CCAA4}3.921 & \cellcolor[HTML]{FFFFFF}1.690 & \cellcolor[HTML]{FFFFFF}1.286 \\
\multicolumn{2}{l}{\fnframe{Event}} &  &  &  &  \\
\textbf{} & nonverbal & \cellcolor[HTML]{FFFFFF}1.431 & \cellcolor[HTML]{FFFFFF}1.503 & \cellcolor[HTML]{FFFFFF}1.186 & \cellcolor[HTML]{FFFFFF}2.339 \\
\textit{} & impersonal & \cellcolor[HTML]{FFFFFF}1.169 & \cellcolor[HTML]{FFFFFF}2.201 & \cellcolor[HTML]{FFFFFF}1.309 & \cellcolor[HTML]{FFFFFF}1.949 \\
\multicolumn{2}{l}{\fnframe{Killing}} & \multicolumn{1}{l}{} & \multicolumn{1}{l}{} & \multicolumn{1}{l}{} & \multicolumn{1}{l}{} \\
\textit{} & nonverbal & \cellcolor[HTML]{FFFFFF}2.007 & \cellcolor[HTML]{FFFFFF}2.387 & \cellcolor[HTML]{FFFFFF}1.032 & \cellcolor[HTML]{FFFFFF}1.673 \\
\textit{} & other & \cellcolor[HTML]{FFFFFF}2.410 & \cellcolor[HTML]{FFFFFF}2.345 & \cellcolor[HTML]{FFFFFF}1.198 & \cellcolor[HTML]{FFFFFF}1.663 \\
\textit{} & active & \cellcolor[HTML]{7DCAA4}3.897 & \cellcolor[HTML]{A6DBC1}2.659 & \cellcolor[HTML]{FFFFFF}1.570 & \cellcolor[HTML]{FFFFFF}1.651 \\
\textit{} & passive & \cellcolor[HTML]{FFFFFF}1.947 & \cellcolor[HTML]{8CD1AF}3.425 & \cellcolor[HTML]{FFFFFF}1.491 & \cellcolor[HTML]{FFFFFF}1.315 \\ \bottomrule
\end{tabular}%
}
\caption{Mean perception scores for ``the main focus is on X''. `*' = differences between frame-construction pairs are significant at $\alpha=0.05$, `**' = significant at $\alpha=0.001$ (Kruskal-Wallis non-parametric H-test). Cells with a value $>2.5$ are highlighted in green.}\label{tab:questionnaire-results}
\end{table}

\section{Perception score prediction}
\label{sec:prediction}

In this section, we introduce models for automatically predicting femicide perception scores, as well as a suite of evaluation measures for evaluating these models. We model our task as a multi-output regression task: given a sentence $S$, we want to predict a perception vector $\vec{p}$, in which every entry $p_i$ represents the value of a particular Likert dimension from the questionnaire (e.g. \textit{`blame on the victim'}, \textit{`focus on an object'}). 

\subsection{Participant aggregation}
In order to train a single model that generalizes over individual participants, we first z-score the perception values for each sentence and each participant and then take the average value across participants. Z-scores are calculated separately for each Likert dimension and participant to account for two types of variability: i) \textit{within-dimension score intensity preference} and ii) \textit{between-dimension preference}. Type (i) refers to different participants making different use of the score range: depending on confidence levels and other factors, participants might choose to make heavy use of the extremities of the range (e.g. very often assign `0' or `5') or concentrate most of their scores in a particular part of the range (i.e. around the center or near the high or low end). Type (ii) refers to the possibility of participants having a tendency to always assign higher or lower scores to particular dimensions. For example, some participants may always give a higher score to `blame on the murderer' vs. `blame on the victim'. By performing regression towards z-scored perception values, we force our models to predict \textit{between-sentence variability}: we are most interested in predicting how each sentence is perceived relative to other sentences (e.g., does this sentence put above-average blame on the victim? below-average focus on the murderer?) and less in absolute scores since these are highly subjective and depend on many individual biases. 

\subsection{Metrics}
We evaluate our multi-output regression problem from several angles. First, we use \textbf{\textit{Root Mean Squared Error (RMSE)}} to measure error rates. This is complemented by \textbf{\textit{R$^\mathbf{2}$}}, which estimates the proportion of variation in the perception scores that is explained by the regression models. R$^2$ is defined both for each dimension and as an average over dimensions. Next, \textbf{\textit{Cosine (COS)}} measures the cosine similarity between the gold and predicted vectors of perception values and provides an estimate of how well the relations between the dimensions are preserved in the mapping. 

An alternative interpretation is the \textbf{\textit{Most Salient Attribute (MSA)}} metric: we evaluate regression as accuracy on the classification task of predicting which Likert dimension has the highest (z-scored) perception value for each question (implemented as simply computing $\mathrm{argmax}$ over the output dimensions corresponding to each question). For example, if for a particular sentence, ``concept'' is the highest-scoring dimension for the \textit{blame} question, this means that ``blame on a concept'' is more salient in this sentence compared to other sentences. Note that the fact that z-scores were computed individually for each dimension makes a major difference here: the dimension with the highest z-scored value does not necessarily also have the highest absolute value. Similarly to the risks of assigning higher or lower scores to particular dimensions, in this case participants may give more points to ``murderer'' on the \textit{blame} question than to ``concept'', even in sentences where ``concept'' is very salient.
In such cases, ``concept'' would always have a lower absolute value than ``murderer'', but might have have a higher z-scored value in sentences where a relatively high score was given to ``concept'' and a relatively low one to ``murderer''.

\subsection{Models}

We compare two types of models: \textit{ridge regression} models (a type of linear regression with L2 regularization) trained on different types of input features, and a selection of relevant pre-trained \textit{transformer} models, fine-tuned for multi-output regression. For reference, we also run a `dummy' baseline model that always predicts the training set mean for each variable.\footnote{\url{https://scikit-learn.org/stable/modules/generated/sklearn.dummy.DummyRegressor.html}}

\paragraph{Features} For the ridge models, we use a series of feature representations with increasing levels of richness. By comparing models trained on different representations, we gain insights into what kind of information is useful for predicting (different aspects of) perception scores. Features are divided into three categories: \textit{Surface} features represent the lexical content of the input sentences, either with simple (unigram) \textit{\textbf{bag-of-words (bow)}} vectors, or with pre-trained \textit{\textbf{FastText (ft)}} embeddings~\cite{grave2018learning}.\footnote{\rebuttal{We chose FastText over competing static embedding models because of its ability to handle out-of-vocabulary tokens. Sentence-level representations were computed by taking the mean over all unigram vectors in the sentence, weighted by occurrence count (i.e., if a word occurs several times in a sentence, it will have a higher weight).}} By contrast, \textit{Frames} features are based on the frame semantic parses of the sentence. The first variant, \textbf{\textit{f1}}, is similar to a bag-of-words, but using counts of any frame instances (e.g. \textit{frm:Commerce\_buy}) and semantic role instances (e.g. \textit{rol:Commerce\_buy:Seller}) present in the sentence instead of unigram counts. Variant \textbf{\textit{f2}} is similar but includes only mentions of our pre-defined frames-of-interest (\fnframe{Killing}, \fnframe{Death}, \ldots). Moreover, \textbf{\textit{f1+}} and \textbf{\textit{f2+}} are versions of \textit{f1} and \textit{f2} that concatenate the bag-of-frame features to the unigram features from \textbf{\textit{bow}}. Finally, \textit{Sentence} features are transformer-derived sentence-level representations. \textit{\textbf{SentenceBERT (sb)}} \cite{reimers-gurevych-2019-sentence} uses representations derived from XLM-R~\cite{conneau2020unsupervised};\footnote{We used pre-trained SentenceBERT models available from \url{https://www.sbert.net/}.} \textit{\textbf{BERT-IT Mean (bm)}} and \textit{\textbf{XLM-R Mean (xm)}} use last-layer representations, averaged over tokens, from Italian BERT XXL and XLM-R, respectively. 

\paragraph{Transformers}
We also implement a neural regression model that consists of a simple linear layer on top of a pre-trained transformer encoder.\footnote{Huggingface Hub links to the exact models used are provided in the Appendix.} We experiment with several variants of BERT with different pretraining corpora and model sizes. \textbf{\textit{Italian BERT XXL Base (BERT-IT)}} is a base-size monolingual BERT model trained on the Italian Wikipedia and the OPUS corpus; \textbf{\textit{BERTino}} is a distilled version of this model. We compare these with \textbf{\textit{Multilingual BERT Base}} \cite{devlin-etal-2019-bert} and \textbf{\textit{Multilingual DistilBERT}} \cite{sanh-distilbert}, trained on concatenated Wikipedia dumps for 104 language, and \textbf{\textit{XLM-RoBERTa Base}} \cite{conneau2020unsupervised}, trained on CommonCrawl data for 100 languages. We use cased models in all cases.

\paragraph{Implementation} Ridge regression models were implemented using \textit{scikit-learn} \cite{scikit-learn}. Transformer models were implemented using \textit{Huggingface Transformers} \cite{huggingface}. We split the dataset into 75\% training and 25\% test data. \rebuttal{We used 6-fold cross-validation within the training set to search for hyperparameters (i.e., six models were trained for each possible setup): $\alpha$ for ridge regression; initial Adam learning rate and weight decay for transformers. The parameters with best performance across folds were then used for training the final model.}

\subsection{Results}

\begin{table*}[]
\centering
\resizebox{\textwidth}{!}{%
\begin{tabular}{lll|r|rrrrrrrrr|rrrrr}
\hline
\textbf{} & \textit{\textbf{}} & \textbf{model} & \multicolumn{1}{c|}{\textit{\textbf{baseline}}} & \multicolumn{9}{c|}{\textit{\textbf{ridge}}} & \multicolumn{5}{c}{\textit{\textbf{transformer}}} \\
 & \textit{\textbf{}} & \textbf{features} & \multicolumn{1}{c|}{\textit{\textbf{}}} & \multicolumn{2}{c|}{\textit{\textbf{Surface}}} & \multicolumn{4}{c|}{\textit{\textbf{Frames}}} & \multicolumn{3}{c|}{\textit{\textbf{Neural}}} & \multicolumn{2}{c|}{\textit{\textbf{bert-it}}} & \multicolumn{2}{c|}{\textit{\textbf{mbert}}} & \multicolumn{1}{c}{\textit{\textbf{xlmr}}} \\
 & \textit{} & \textit{} & \multicolumn{1}{l|}{\textbf{}} & \multicolumn{1}{c}{\textit{\textbf{bow}}} & \multicolumn{1}{c|}{\textit{\textbf{ft}}} & \multicolumn{1}{l}{\textit{\textbf{f1}}} & \multicolumn{1}{l}{\textit{\textbf{f1+}}} & \multicolumn{1}{l}{\textit{\textbf{f2}}} & \multicolumn{1}{l|}{\textit{\textbf{f2+}}} & \multicolumn{1}{l}{\textit{\textbf{sb}}} & \multicolumn{1}{l}{\textit{\textbf{bm}}} & \multicolumn{1}{l|}{\textit{\textbf{xm}}} & \multicolumn{1}{l}{\textbf{base}} & \multicolumn{1}{l|}{\textbf{dist}} & \multicolumn{1}{l}{\textbf{base}} & \multicolumn{1}{l|}{\textbf{dist}} & \multicolumn{1}{l}{\textbf{base}} \\ \hline
\multicolumn{2}{l}{\textbf{RMSE}} & \textit{} & \cellcolor[HTML]{EFA7A1}0.67 & \cellcolor[HTML]{F1B3AE}0.59 & \multicolumn{1}{r|}{\cellcolor[HTML]{F1B1AC}0.60} & \cellcolor[HTML]{F1B2AD}0.59 & \cellcolor[HTML]{F1B4AF}0.58 & \cellcolor[HTML]{F0ACA7}0.63 & \multicolumn{1}{r|}{\cellcolor[HTML]{F1B2AC}0.59} & \cellcolor[HTML]{F2B6B1}0.56 & \cellcolor[HTML]{F2B9B4}0.54 & \cellcolor[HTML]{F2B7B2}0.56 & \cellcolor[HTML]{F4C1BD}0.48 & \multicolumn{1}{r|}{\cellcolor[HTML]{F4C2BD}0.47} & \cellcolor[HTML]{F3BDB8}0.51 & \multicolumn{1}{r|}{\cellcolor[HTML]{F2BAB5}0.53} & \cellcolor[HTML]{F3BDB9}0.51 \\ \hline
\multicolumn{2}{l}{\textbf{COS}} &  & \cellcolor[HTML]{FFFFFF}-0.02 & \cellcolor[HTML]{AEDEC6}0.49 & \multicolumn{1}{r|}{\cellcolor[HTML]{B2E0CA}0.46} & \cellcolor[HTML]{AFDFC8}0.48 & \cellcolor[HTML]{A8DCC3}0.52 & \cellcolor[HTML]{C3E7D5}0.36 & \multicolumn{1}{r|}{\cellcolor[HTML]{ACDEC5}0.50} & \cellcolor[HTML]{A4DABF}0.55 & \cellcolor[HTML]{9DD8BB}0.58 & \cellcolor[HTML]{A3DABF}0.55 & \cellcolor[HTML]{8FD2B1}0.67 & \multicolumn{1}{r|}{\cellcolor[HTML]{8CD1AF}0.69} & \cellcolor[HTML]{93D4B4}0.65 & \multicolumn{1}{r|}{\cellcolor[HTML]{9FD8BC}0.58} & \cellcolor[HTML]{90D2B2}0.66 \\ \hline
 & \textit{\textbf{Average}} & \textit{\textbf{}} & \cellcolor[HTML]{FFFFFF}-0.01 & \cellcolor[HTML]{DDF2E8}0.20 & \multicolumn{1}{r|}{\cellcolor[HTML]{DFF2E9}0.19} & \cellcolor[HTML]{DEF2E8}0.20 & \cellcolor[HTML]{D9F0E5}0.23 & \cellcolor[HTML]{F2FAF6}0.08 & \multicolumn{1}{r|}{\cellcolor[HTML]{E1F3EA}0.18} & \cellcolor[HTML]{D1EDDF}0.28 & \cellcolor[HTML]{C9E9D9}0.33 & \cellcolor[HTML]{D0ECDF}0.28 & \cellcolor[HTML]{B5E1CC}0.44 & \multicolumn{1}{r|}{\cellcolor[HTML]{B4E1CB}0.45} & \cellcolor[HTML]{C0E6D3}0.38 & \multicolumn{1}{r|}{\cellcolor[HTML]{C7E8D8}0.34} & \cellcolor[HTML]{C0E6D3}0.38 \\ \cline{2-18} 
 &  & \textit{\textbf{murderer}} & \cellcolor[HTML]{FFFFFF}0.00 & \cellcolor[HTML]{AEDEC7}0.49 & \multicolumn{1}{r|}{\cellcolor[HTML]{D0ECDF}0.28} & \cellcolor[HTML]{CEEBDD}0.30 & \cellcolor[HTML]{C3E7D5}0.36 & \cellcolor[HTML]{ECF8F2}0.11 & \multicolumn{1}{r|}{\cellcolor[HTML]{C2E6D4}0.37} & \cellcolor[HTML]{AFDFC7}0.48 & \cellcolor[HTML]{B5E1CC}0.44 & \cellcolor[HTML]{B2E0C9}0.46 & \cellcolor[HTML]{A1D9BE}0.56 & \multicolumn{1}{r|}{\cellcolor[HTML]{99D6B8}0.61} & \cellcolor[HTML]{A9DDC3}0.51 & \multicolumn{1}{r|}{\cellcolor[HTML]{B0DFC8}0.47} & \cellcolor[HTML]{ACDEC5}0.50 \\
 &  & \textit{\textbf{victim}} & \cellcolor[HTML]{FFFFFF}0.00 & \cellcolor[HTML]{FFFFFF}-0.05 & \multicolumn{1}{r|}{\cellcolor[HTML]{FFFFFF}-0.01} & \cellcolor[HTML]{FFFFFF}-0.03 & \cellcolor[HTML]{FFFFFF}-0.03 & \cellcolor[HTML]{FFFFFF}0.00 & \multicolumn{1}{r|}{\cellcolor[HTML]{FFFFFF}-0.08} & \cellcolor[HTML]{F1FAF6}0.09 & \cellcolor[HTML]{E9F7F0}0.13 & \cellcolor[HTML]{F1F9F5}0.09 & \cellcolor[HTML]{E3F4EC}0.17 & \multicolumn{1}{r|}{\cellcolor[HTML]{D8EFE4}0.24} & \cellcolor[HTML]{E7F6EE}0.15 & \multicolumn{1}{r|}{\cellcolor[HTML]{FEFFFE}0.01} & \cellcolor[HTML]{EFF9F4}0.10 \\
 &  & \textit{\textbf{concept}} & \cellcolor[HTML]{FFFFFF}0.00 & \cellcolor[HTML]{F8FCFA}0.05 & \multicolumn{1}{r|}{\cellcolor[HTML]{ECF8F2}0.11} & \cellcolor[HTML]{F3FAF7}0.08 & \cellcolor[HTML]{F3FBF7}0.07 & \cellcolor[HTML]{FDFEFD}0.02 & \multicolumn{1}{r|}{\cellcolor[HTML]{F0F9F5}0.09} & \cellcolor[HTML]{DBF1E6}0.22 & \cellcolor[HTML]{D1EDDF}0.27 & \cellcolor[HTML]{D8F0E4}0.23 & \cellcolor[HTML]{C2E7D5}0.37 & \multicolumn{1}{r|}{\cellcolor[HTML]{C8E9D9}0.33} & \cellcolor[HTML]{D5EEE2}0.26 & \multicolumn{1}{r|}{\cellcolor[HTML]{EBF7F1}0.12} & \cellcolor[HTML]{D6EFE2}0.25 \\
 &  & \textit{\textbf{object}} & \cellcolor[HTML]{FFFFFF}0.00 & \cellcolor[HTML]{F6FBF9}0.06 & \multicolumn{1}{r|}{\cellcolor[HTML]{EAF7F0}0.13} & \cellcolor[HTML]{EEF8F3}0.11 & \cellcolor[HTML]{EDF8F3}0.11 & \cellcolor[HTML]{EEF8F3}0.11 & \multicolumn{1}{r|}{\cellcolor[HTML]{F9FDFB}0.04} & \cellcolor[HTML]{E8F6EF}0.14 & \cellcolor[HTML]{E1F3EA}0.18 & \cellcolor[HTML]{E8F6EF}0.14 & \cellcolor[HTML]{D5EEE2}0.25 & \multicolumn{1}{r|}{\cellcolor[HTML]{CBEADB}0.31} & \cellcolor[HTML]{DBF1E6}0.22 & \multicolumn{1}{r|}{\cellcolor[HTML]{D5EEE2}0.25} & \cellcolor[HTML]{DDF2E8}0.20 \\
 & \multirow{-5}{*}{\textit{\textbf{Blame}}} & \textit{\textbf{no-one}} & \cellcolor[HTML]{FFFFFF}-0.02 & \cellcolor[HTML]{CAEADA}0.32 & \multicolumn{1}{r|}{\cellcolor[HTML]{E1F3EA}0.18} & \cellcolor[HTML]{DDF2E7}0.21 & \cellcolor[HTML]{D7EFE3}0.24 & \cellcolor[HTML]{FDFEFE}0.02 & \multicolumn{1}{r|}{\cellcolor[HTML]{D0ECDE}0.28} & \cellcolor[HTML]{C1E6D4}0.37 & \cellcolor[HTML]{D0ECDE}0.28 & \cellcolor[HTML]{BDE5D1}0.39 & \cellcolor[HTML]{BFE5D2}0.39 & \multicolumn{1}{r|}{\cellcolor[HTML]{C8E9D9}0.33} & \cellcolor[HTML]{BCE4D1}0.40 & \multicolumn{1}{r|}{\cellcolor[HTML]{C6E8D7}0.34} & \cellcolor[HTML]{CFECDD}0.29 \\ \cline{2-18} 
 &  & \textit{\textbf{human}} & \cellcolor[HTML]{FFFFFF}-0.01 & \cellcolor[HTML]{C0E6D3}0.38 & \multicolumn{1}{r|}{\cellcolor[HTML]{D0ECDE}0.28} & \cellcolor[HTML]{D2EDE0}0.27 & \cellcolor[HTML]{C5E8D7}0.35 & \cellcolor[HTML]{EAF7F1}0.13 & \multicolumn{1}{r|}{\cellcolor[HTML]{CBEADB}0.31} & \cellcolor[HTML]{ABDDC4}0.50 & \cellcolor[HTML]{C2E7D5}0.37 & \cellcolor[HTML]{C2E7D5}0.37 & \cellcolor[HTML]{A2DABE}0.56 & \multicolumn{1}{r|}{\cellcolor[HTML]{9AD7B9}0.60} & \cellcolor[HTML]{AADDC4}0.51 & \multicolumn{1}{r|}{\cellcolor[HTML]{AEDEC7}0.49} & \cellcolor[HTML]{BAE3CF}0.41 \\
 &  & \textit{\textbf{object}} & \cellcolor[HTML]{FFFFFF}0.00 & \cellcolor[HTML]{B5E1CB}0.45 & \multicolumn{1}{r|}{\cellcolor[HTML]{CBEADB}0.31} & \cellcolor[HTML]{AADDC4}0.51 & \cellcolor[HTML]{A4DABF}0.55 & \cellcolor[HTML]{BDE5D1}0.40 & \multicolumn{1}{r|}{\cellcolor[HTML]{AADDC4}0.51} & \cellcolor[HTML]{C4E7D6}0.35 & \cellcolor[HTML]{A5DBC1}0.54 & \cellcolor[HTML]{B5E1CC}0.44 & \cellcolor[HTML]{78C9A1}0.80 & \multicolumn{1}{r|}{\cellcolor[HTML]{78C9A1}0.81} & \cellcolor[HTML]{7CCAA4}0.79 & \multicolumn{1}{r|}{\cellcolor[HTML]{8ED2B0}0.68} & \cellcolor[HTML]{83CDA9}0.74 \\
 &  & \textit{\textbf{no-one}} & \cellcolor[HTML]{FFFFFF}-0.01 & \cellcolor[HTML]{FFFFFF}-0.16 & \multicolumn{1}{r|}{\cellcolor[HTML]{F0F9F5}0.09} & \cellcolor[HTML]{FFFFFF}-0.05 & \cellcolor[HTML]{FFFFFF}-0.11 & \cellcolor[HTML]{FFFFFF}-0.18 & \multicolumn{1}{r|}{\cellcolor[HTML]{FFFFFF}-0.23} & \cellcolor[HTML]{FFFFFF}-0.22 & \cellcolor[HTML]{FBFEFC}0.03 & \cellcolor[HTML]{FFFFFF}-0.12 & \cellcolor[HTML]{FFFFFF}-0.07 & \multicolumn{1}{r|}{\cellcolor[HTML]{FFFFFF}-0.07} & \cellcolor[HTML]{EFF9F4}0.10 & \multicolumn{1}{r|}{\cellcolor[HTML]{EDF8F3}0.11} & \cellcolor[HTML]{FFFFFF}-0.09 \\
 & \multirow{-4}{*}{\textit{\textbf{Cause}}} & \textit{\textbf{concept}} & \cellcolor[HTML]{FFFFFF}-0.01 & \cellcolor[HTML]{EEF8F3}0.11 & \multicolumn{1}{r|}{\cellcolor[HTML]{FBFEFC}0.03} & \cellcolor[HTML]{DFF2E9}0.20 & \cellcolor[HTML]{DFF2E9}0.19 & \cellcolor[HTML]{FFFFFF}-0.02 & \multicolumn{1}{r|}{\cellcolor[HTML]{EEF8F3}0.11} & \cellcolor[HTML]{F3FBF7}0.07 & \cellcolor[HTML]{E0F3EA}0.19 & \cellcolor[HTML]{FFFFFF}0.00 & \cellcolor[HTML]{BEE5D2}0.39 & \multicolumn{1}{r|}{\cellcolor[HTML]{CBEADB}0.31} & \cellcolor[HTML]{F8FCFA}0.04 & \multicolumn{1}{r|}{\cellcolor[HTML]{E1F3EA}0.18} & \cellcolor[HTML]{CCEBDC}0.31 \\ \cline{2-18} 
 &  & \textit{\textbf{murderer}} & \cellcolor[HTML]{FFFFFF}-0.01 & \cellcolor[HTML]{AADDC4}0.51 & \multicolumn{1}{r|}{\cellcolor[HTML]{C7E9D8}0.33} & \cellcolor[HTML]{C6E8D8}0.34 & \cellcolor[HTML]{B7E2CD}0.43 & \cellcolor[HTML]{E7F6EE}0.15 & \multicolumn{1}{r|}{\cellcolor[HTML]{B8E3CE}0.42} & \cellcolor[HTML]{AEDFC7}0.48 & \cellcolor[HTML]{B7E2CD}0.43 & \cellcolor[HTML]{AADDC4}0.51 & \cellcolor[HTML]{91D3B3}0.66 & \multicolumn{1}{r|}{\cellcolor[HTML]{92D3B3}0.65} & \cellcolor[HTML]{99D6B8}0.61 & \multicolumn{1}{r|}{\cellcolor[HTML]{9AD6B9}0.61} & \cellcolor[HTML]{9DD8BB}0.58 \\
 &  & \textit{\textbf{victim}} & \cellcolor[HTML]{FFFFFF}-0.03 & \cellcolor[HTML]{C8E9D9}0.33 & \multicolumn{1}{r|}{\cellcolor[HTML]{CFECDE}0.29} & \cellcolor[HTML]{D4EEE1}0.26 & \cellcolor[HTML]{C8E9D9}0.33 & \cellcolor[HTML]{DEF2E8}0.20 & \multicolumn{1}{r|}{\cellcolor[HTML]{CBEADB}0.31} & \cellcolor[HTML]{AEDEC7}0.49 & \cellcolor[HTML]{A1D9BE}0.56 & \cellcolor[HTML]{AFDFC8}0.48 & \cellcolor[HTML]{9DD7BB}0.59 & \multicolumn{1}{r|}{\cellcolor[HTML]{96D5B6}0.63} & \cellcolor[HTML]{ADDEC6}0.49 & \multicolumn{1}{r|}{\cellcolor[HTML]{AFDFC8}0.48} & \cellcolor[HTML]{99D6B8}0.61 \\
 &  & \textit{\textbf{concept}} & \cellcolor[HTML]{FFFFFF}0.00 & \cellcolor[HTML]{EAF7F0}0.13 & \multicolumn{1}{r|}{\cellcolor[HTML]{D1EDDF}0.28} & \cellcolor[HTML]{CCEADB}0.31 & \cellcolor[HTML]{CAEADA}0.32 & \cellcolor[HTML]{F0F9F5}0.09 & \multicolumn{1}{r|}{\cellcolor[HTML]{E5F5ED}0.16} & \cellcolor[HTML]{CEEBDD}0.30 & \cellcolor[HTML]{B1E0C9}0.47 & \cellcolor[HTML]{D5EEE2}0.25 & \cellcolor[HTML]{96D5B6}0.63 & \multicolumn{1}{r|}{\cellcolor[HTML]{94D4B5}0.64} & \cellcolor[HTML]{95D4B5}0.64 & \multicolumn{1}{r|}{\cellcolor[HTML]{B3E0CA}0.46} & \cellcolor[HTML]{95D4B5}0.64 \\
\multirow{-14}{*}{\textbf{R$^2$}} & \multirow{-4}{*}{\textit{\textbf{Focus}}} & \textit{\textbf{object}} & \cellcolor[HTML]{FFFFFF}0.00 & \cellcolor[HTML]{F5FBF8}0.06 & \multicolumn{1}{r|}{\cellcolor[HTML]{DDF2E7}0.21} & \cellcolor[HTML]{E7F6EF}0.14 & \cellcolor[HTML]{EAF7F1}0.13 & \cellcolor[HTML]{F3FAF7}0.08 & \multicolumn{1}{r|}{\cellcolor[HTML]{F4FBF7}0.07} & \cellcolor[HTML]{C9EADA}0.32 & \cellcolor[HTML]{C3E7D5}0.36 & \cellcolor[HTML]{BAE3CF}0.41 & \cellcolor[HTML]{B2E0CA}0.46 & \multicolumn{1}{r|}{\cellcolor[HTML]{B3E0CA}0.46} & \cellcolor[HTML]{E0F3EA}0.19 & \multicolumn{1}{r|}{\cellcolor[HTML]{DDF1E7}0.21} & \cellcolor[HTML]{C2E6D4}0.37 \\ \hline
\end{tabular}%
}
\caption{Regression results overview: RMSE, Cosine Similarity, and R$^2$ scores}\label{tab:results-overview}
\end{table*}
\begin{table*}[]
\centering
\resizebox{.9\textwidth}{!}{%
\begin{tabular}{l|r|rrrrrrrrr|rrrrr}
\hline
\textbf{model} & \multicolumn{1}{c|}{\textit{\textbf{baseline}}} & \multicolumn{9}{c|}{\textit{\textbf{ridge}}} & \multicolumn{5}{c}{\textit{\textbf{transformer}}} \\
\textbf{features} & \multicolumn{1}{c|}{\textit{\textbf{}}} & \multicolumn{2}{c|}{\textit{\textbf{Surface}}} & \multicolumn{4}{c|}{\textit{\textbf{Frames}}} & \multicolumn{3}{c|}{\textit{\textbf{Neural}}} & \multicolumn{2}{c|}{\textit{\textbf{bert-it}}} & \multicolumn{2}{c|}{\textit{\textbf{mbert}}} & \multicolumn{1}{c}{\textit{\textbf{xlmr}}} \\
\textit{} & \multicolumn{1}{l|}{\textbf{}} & \multicolumn{1}{c}{\textit{\textbf{bow}}} & \multicolumn{1}{c|}{\textit{\textbf{ft}}} & \multicolumn{1}{c}{\textit{\textbf{f1}}} & \multicolumn{1}{c}{\textit{\textbf{f1+}}} & \multicolumn{1}{c}{\textit{\textbf{f2}}} & \multicolumn{1}{c|}{\textit{\textbf{f2+}}} & \multicolumn{1}{c}{\textit{\textbf{sb}}} & \multicolumn{1}{c}{\textit{\textbf{bm}}} & \multicolumn{1}{c|}{\textit{\textbf{xm}}} & \multicolumn{1}{c}{\textbf{base}} & \multicolumn{1}{c|}{\textbf{dist}} & \multicolumn{1}{c}{\textbf{base}} & \multicolumn{1}{c|}{\textbf{dist}} & \multicolumn{1}{c}{\textbf{base}} \\ \hline
\textit{\textbf{Blame}} & \cellcolor[HTML]{D4EEE1}0.26 & \cellcolor[HTML]{B6E2CC}0.44 & \multicolumn{1}{r|}{\cellcolor[HTML]{B2E0CA}0.46} & \cellcolor[HTML]{B6E2CC}0.44 & \cellcolor[HTML]{B1E0C9}0.47 & \cellcolor[HTML]{BEE5D2}0.39 & \multicolumn{1}{r|}{\cellcolor[HTML]{B2E0CA}0.46} & \cellcolor[HTML]{ADDEC6}0.49 & \cellcolor[HTML]{A8DCC3}0.52 & \cellcolor[HTML]{B1E0C9}0.47 & \cellcolor[HTML]{ABDDC5}0.50 & \multicolumn{1}{r|}{\cellcolor[HTML]{A1D9BE}0.56} & \cellcolor[HTML]{AADDC4}0.51 & \multicolumn{1}{r|}{\cellcolor[HTML]{B1E0C9}0.47} & \cellcolor[HTML]{A6DBC1}0.53 \\
\textit{\textbf{Cause}} & \cellcolor[HTML]{D2EDE0}0.27 & \cellcolor[HTML]{B4E1CB}0.45 & \multicolumn{1}{r|}{\cellcolor[HTML]{ADDEC6}0.49} & \cellcolor[HTML]{ADDEC6}0.49 & \cellcolor[HTML]{A3DABF}0.55 & \cellcolor[HTML]{B4E1CB}0.45 & \multicolumn{1}{r|}{\cellcolor[HTML]{A3DABF}0.55} & \cellcolor[HTML]{B2E0CA}0.46 & \cellcolor[HTML]{A8DCC3}0.52 & \cellcolor[HTML]{A1D9BE}0.56 & \cellcolor[HTML]{94D4B5}0.64 & \multicolumn{1}{r|}{\cellcolor[HTML]{8FD2B1}0.67} & \cellcolor[HTML]{9CD7BA}0.59 & \multicolumn{1}{r|}{\cellcolor[HTML]{A0D9BD}0.57} & \cellcolor[HTML]{9BD7B9}0.60 \\
\textit{\textbf{Focus}} & \cellcolor[HTML]{D7EFE3}0.24 & \cellcolor[HTML]{A1D9BE}0.56 & \multicolumn{1}{r|}{\cellcolor[HTML]{96D5B6}0.63} & \cellcolor[HTML]{ADDEC6}0.49 & \cellcolor[HTML]{A0D9BD}0.57 & \cellcolor[HTML]{B9E3CE}0.42 & \multicolumn{1}{r|}{\cellcolor[HTML]{A0D9BD}0.57} & \cellcolor[HTML]{97D5B7}0.62 & \cellcolor[HTML]{97D5B7}0.62 & \cellcolor[HTML]{9BD7B9}0.60 & \cellcolor[HTML]{85CEAA}0.73 & \multicolumn{1}{r|}{\cellcolor[HTML]{87CFAB}0.72} & \cellcolor[HTML]{97D5B7}0.62 & \multicolumn{1}{r|}{\cellcolor[HTML]{A0D9BD}0.57} & \cellcolor[HTML]{8AD0AE}0.70 \\ \hline
\textit{\textbf{mean}} & \cellcolor[HTML]{D4EEE1}0.26 & \cellcolor[HTML]{AEDFC7}0.48 & \multicolumn{1}{r|}{\cellcolor[HTML]{A7DCC2}0.53} & \cellcolor[HTML]{B0DFC8}0.47 & \cellcolor[HTML]{A6DBC1}0.53 & \cellcolor[HTML]{B9E3CE}0.42 & \multicolumn{1}{r|}{\cellcolor[HTML]{A7DCC2}0.53} & \cellcolor[HTML]{A8DCC2}0.52 & \cellcolor[HTML]{A3DABF}0.55 & \cellcolor[HTML]{A4DBC0}0.54 & \cellcolor[HTML]{97D5B7}0.62 & \multicolumn{1}{r|}{\cellcolor[HTML]{92D3B3}0.65} & \cellcolor[HTML]{A0D9BD}0.57 & \multicolumn{1}{r|}{\cellcolor[HTML]{A5DBC1}0.54} & \cellcolor[HTML]{99D6B8}0.61 \\ \hline
\end{tabular}%
}
\caption{Most Salient Attribute scores}\label{tab:maxcol}
\end{table*}

Table~\ref{tab:results-overview} shows the main results on the \rebuttal{test} set for the RMSE, COS and R$^2$ metrics. Strongest results are obtained with the fine-tuned monolingual BERT models across all measures, with an overall R$^2$ scores around 0.45, meaning that these models explain almost half of the observed variance in perception scores. The multilingual BERT models (mBERT and XLM-R) perform consistently worse, with an average R$^2$ of 0.38 or below. Interestingly, we observe a drop in performance between the full-size and distilled models for mBERT, but not for the monolingual Italian BERT, where BERTino even performs slightly better than the original model. Drops in R$^2$ do not always align with drops in cosine scores: for example, XLM-R scores 0.06 R$^2$ points lower than BERT-IT/base, but the cosine score drops by only 0.01, while mBERT/dist loses 0.10 points on R$^2$ and 0.09 on COS. Thus, it appears that some models (like XLM-R) are less accurate at predicting the exact magnitude of perception scores but relatively good at capturing the overall score pattern across dimensions.  

While the ridge regression models perform substantially worse than the transformer models, comparing the results between different feature representations is insightful for understanding what information is needed to predict perception: the Surface and Frames models all perform similarly with R$^2$ scores around 0.20 (with \textit{f2} as a negative outlier), while the models with Neural features perform better (R$^2$ 0.28-0.33). Simple counts of unigrams (\textit{bow}) and frames (\textit{f1}) give very similar overall scores; concatenating these features (\textit{f1+}) leads to a small improvement (+0.03 R$^2$). This suggests that frames are useful for summarizing relevant lexical material (grouping together lexical units), but that the additional information about semantic and syntactic structure that is provided by role and construction labels does not lead to substantial gains. Using FastText embeddings instead of unigrams does not lead to gains, either. Meanwhile, comparing ridge models trained on transformer-derived features, we find best results with mean last layer representations from Italian BERT (\textit{bm}), with slightly lower scores for the two models based on XLM-R (\textit{sb} and \textit{xm}); surprisingly, SentenceBERT (\textit{sb}) does not seem to have an advantage over averaged last-layer representations (\textit{xm}). \\
\indent Comparing R$^2$ scores across different questions and attributes reveals large differences in difficulty of prediction: for example, \textit{blame on murderer} gets good scores across models, while \textit{blame on victim} has relatively poor scores even for the strongest models (e.g. 0.24 for BERTino), and at-baseline (or worse) scores for the weaker models --- notably, distilled mBERT, which performs decently on other attributes. \textit{Caused by no-one} is even harder to predict, with no model scoring above 0.10. The \textit{Focus} question has the overall best and most consistent performance, especially for the Italian BERT-based models, which achieve decent performance (0.46-0.66 R$^2$) for each of the four attributes.

\begin{table*}[]
\centering
\resizebox{\textwidth}{!}{%
\begin{tabular}{@{}llclclcllclclcc@{}}
\toprule
 & \multicolumn{6}{c}{\textbf{blame: murderer}} &  & \multicolumn{6}{c}{\textbf{focus: concept}} \\
 & \multicolumn{2}{c}{\textbf{ridge/bow}} & \multicolumn{2}{c}{\textbf{ridge/f1+}} & \multicolumn{2}{c}{\textbf{bertino}} &  & \multicolumn{2}{c}{\textbf{ridge/bow}} & \multicolumn{2}{c}{\textbf{ridge/f1+}} & \multicolumn{2}{c}{\textbf{bertino}} \\
 & \textit{feature} & \multicolumn{1}{l}{\textit{attr}} & \textit{feature} & \multicolumn{1}{l}{\textit{attr}} & \textit{feature} & \multicolumn{1}{l}{\textit{attr}} &  & \textit{feature} & \textit{attr} & \textit{feature} & \multicolumn{1}{l}{\textit{attr}} & \textit{feature} & \multicolumn{1}{l}{\textit{attr}} \\ \midrule
+1 & \begin{tabular}[c]{@{}l@{}}ex \\ \textit{\small{[}'ex' (ex-partner){]}}\end{tabular} & \cellcolor[HTML]{57BB8A}\small{0.38} & rol:Killing:Killer & \cellcolor[HTML]{57BB8A}\small{0.21} & \begin{tabular}[c]{@{}l@{}}killer \\ \textit{\small{[}'killer'{]}}\end{tabular} & \cellcolor[HTML]{7CCAA4}\small{0.79} &  & \begin{tabular}[c]{@{}l@{}}che \\ \textit{\small{[}'that' (rel.pn./comp.){]}}\end{tabular} & \cellcolor[HTML]{57BB8A}\small{0.20} & \begin{tabular}[c]{@{}l@{}}che \\ \textit{\small{[}'that' (rpn./cmp.){]}}\end{tabular} & \cellcolor[HTML]{57BB8A}\small{0.12} & \begin{tabular}[c]{@{}l@{}}femminicidio \\ \textit{\small{[}'femicide'{]}}\end{tabular} & \cellcolor[HTML]{A9DDC3}\small{0.49} \\
+2 & \begin{tabular}[c]{@{}l@{}}uccide \\ \textit{\small{[}'he/she/it kills'{]}}\end{tabular} & \cellcolor[HTML]{57BB8A}\small{0.33} & \begin{tabular}[c]{@{}l@{}}ex \\ \textit{\small{[}'ex' (ex-partner){]}}\end{tabular} & \cellcolor[HTML]{57BB8A}\small{0.15} & \begin{tabular}[c]{@{}l@{}}uccide \\ \textit{\small{[}'he/she/it kills'{]}}\end{tabular} & \cellcolor[HTML]{83CDA9}\small{0.75} &  & \begin{tabular}[c]{@{}l@{}}pista\\ \textit{\small{[}'course of events'{]}}\end{tabular} & \cellcolor[HTML]{57BB8A}\small{0.19} & \begin{tabular}[c]{@{}l@{}}sara\\ \textit{\small{[}'he/she/it will be'{]}}\end{tabular} & \cellcolor[HTML]{6EC49A}\small{0.09} & \begin{tabular}[c]{@{}l@{}}figlio\\ \textit{\small{[}'son'{]}}\end{tabular} & \cellcolor[HTML]{C7E9D8}\small{0.31} \\
+3 & \begin{tabular}[c]{@{}l@{}}moglie \\ \textit{\small{[}'wife'{]}}\end{tabular} & \cellcolor[HTML]{57BB8A}\small{0.31} & frm:Pers\_rel & \cellcolor[HTML]{57BB8A}\small{0.14} & \begin{tabular}[c]{@{}l@{}}assassino \\ \textit{\small{[}'murderer'{]}}\end{tabular} & \cellcolor[HTML]{89CFAD}\small{0.71} &  & \begin{tabular}[c]{@{}l@{}}passionale\\ \textit{\small{[}'out of passion'{]}}\end{tabular} & \cellcolor[HTML]{57BB8A}\small{0.19} & \begin{tabular}[c]{@{}l@{}}pista\\ \textit{\small{[}'course of events'{]}}\end{tabular} & \cellcolor[HTML]{79C7A1}\small{0.08} & \begin{tabular}[c]{@{}l@{}}non\\ \textit{\small{[}'not'{]}}\end{tabular} & \cellcolor[HTML]{DCF1E7}\small{0.17} \\
+4 & \begin{tabular}[c]{@{}l@{}}uccise \\ \textit{\small{[}'killed' (ptc, f.pl.){]}}\end{tabular} & \cellcolor[HTML]{57BB8A}\small{0.24} & frm:Killing & \cellcolor[HTML]{57BB8A}\small{0.13} & \begin{tabular}[c]{@{}l@{}}ex \\ \textit{\small{[}'ex' (ex-partner){]}}\end{tabular} & \cellcolor[HTML]{99D6B8}\small{0.62} &  & \begin{tabular}[c]{@{}l@{}}sara\\ \textit{\small{[}'he/she/it will be'{]}}\end{tabular} & \cellcolor[HTML]{57BB8A}\small{0.19} & \begin{tabular}[c]{@{}l@{}}non\\ \textit{\small{[}'not'{]}}\end{tabular} & \cellcolor[HTML]{7DC9A4}\small{0.08} & : & \cellcolor[HTML]{DDF2E8}\small{0.17} \\
+5 & \begin{tabular}[c]{@{}l@{}}assassino \\ \textit{\small{[}'murderer'{]}}\end{tabular} & \cellcolor[HTML]{57BB8A}\small{0.22} & cx:Pers\_rel++nvrb & \cellcolor[HTML]{57BB8A}\small{0.13} & \begin{tabular}[c]{@{}l@{}}fidanzato \\ \textit{\small{[}'boyfriend'{]}}\end{tabular} & \cellcolor[HTML]{ABDDC5}\small{0.51} &  & \begin{tabular}[c]{@{}l@{}}femminicidio\\ \textit{\small{[}'femicide'{]}}\end{tabular} & \cellcolor[HTML]{57BB8A}\small{0.17} & \begin{tabular}[c]{@{}l@{}}femminicidio\\ \textit{\small{[}'femicide'{]}}\end{tabular} & \cellcolor[HTML]{7DC9A4}\small{0.08} & \begin{tabular}[c]{@{}l@{}}suicidio\\ \textit{\small{[}'suicide'{]}}\end{tabular} & \cellcolor[HTML]{DFF2E9}\small{0.15} \\
\rowcolor[HTML]{FFFFFF} 
{\color[HTML]{FFFFFF} } & {\color[HTML]{FFFFFF} } & \multicolumn{1}{l}{\cellcolor[HTML]{FFFFFF}{\color[HTML]{FFFFFF} }} & {\color[HTML]{FFFFFF} } & \multicolumn{1}{l}{\cellcolor[HTML]{FFFFFF}{\color[HTML]{FFFFFF} }} & {\color[HTML]{FFFFFF} } & \multicolumn{1}{l}{\cellcolor[HTML]{FFFFFF}{\color[HTML]{FFFFFF} }} & {\color[HTML]{FFFFFF} } & {\color[HTML]{FFFFFF} } & {\color[HTML]{FFFFFF} } & {\color[HTML]{FFFFFF} } & \multicolumn{1}{l}{\cellcolor[HTML]{FFFFFF}{\color[HTML]{FFFFFF} }} & {\color[HTML]{FFFFFF} } & \multicolumn{1}{l}{\cellcolor[HTML]{FFFFFF}{\color[HTML]{FFFFFF} }} \\
-5 & \begin{tabular}[c]{@{}l@{}}sono \\ \textit{\small{[}'I am' / 'they are'{]}}\end{tabular} & \cellcolor[HTML]{E67C73}\small{-0.14} & rol:Event:Event & \cellcolor[HTML]{EBB0AB}\small{-0.06} & \begin{tabular}[c]{@{}l@{}}una \\ \textit{\small{[}'a' (f.){]}}\end{tabular} & \cellcolor[HTML]{FBEAE9}\small{-0.14} &  & \begin{tabular}[c]{@{}l@{}}omicida\\ \textit{\small{[}'murderer'{]}}\end{tabular} & \cellcolor[HTML]{E67C73}\small{-0.13} & \begin{tabular}[c]{@{}l@{}}nell'\\ \textit{\small{[}'in the'{]}}\end{tabular} & \cellcolor[HTML]{EAA59F}\small{-0.07} & \begin{tabular}[c]{@{}l@{}}uccisa\\ \textit{\small{[}'killed' (ptc, f.sg.){]}}\end{tabular} & \cellcolor[HTML]{F7D9D7}\small{-0.32} \\
-4 & \begin{tabular}[c]{@{}l@{}}della \\ \textit{\small{[}'of the' (+ f.noun){]}}\end{tabular} & \cellcolor[HTML]{E67C73}\small{-0.15} & \begin{tabular}[c]{@{}l@{}}sono \\ \textit{\small{[}'I am' / 'they are'{]}}\end{tabular} & \cellcolor[HTML]{EBADA7}\small{-0.06} & . & \cellcolor[HTML]{FAE9E8}\small{-0.14} &  & \begin{tabular}[c]{@{}l@{}}trovata \\ \textit{\small{[}'found' (ptc, f.sg.){]}}\end{tabular} & \cellcolor[HTML]{E67C73}\small{-0.14} & \begin{tabular}[c]{@{}l@{}}della \\ \textit{\small{[}'of the' (+ f.noun){]}}\end{tabular} & \cellcolor[HTML]{EAA49E}\small{-0.07} & \begin{tabular}[c]{@{}l@{}}morta \\ \textit{\small{[}'dead' (f.sg.){]}}\end{tabular} & \cellcolor[HTML]{F7D9D6}\small{-0.32} \\
-3 & - & \cellcolor[HTML]{E67C73}\small{-0.16} & frm:Event & \cellcolor[HTML]{E99891}\small{-0.08} & \begin{tabular}[c]{@{}l@{}}sono \\ \textit{\small{[}'I am' / 'they are'{]}}\end{tabular} & \cellcolor[HTML]{FAE8E6}\small{-0.15} &  & \begin{tabular}[c]{@{}l@{}}nell'\\ \textit{\small{[}'in the'{]}}\end{tabular} & \cellcolor[HTML]{E67C73}\small{-0.14} & \begin{tabular}[c]{@{}l@{}}due\\ \textit{\small{[}'two'{]}}\end{tabular} & \cellcolor[HTML]{EAA49E}\small{-0.07} & \begin{tabular}[c]{@{}l@{}}killer \\ \textit{\small{[}'killer'{]}}\end{tabular} & \cellcolor[HTML]{F6D0CD}\small{-0.38} \\
-2 & \begin{tabular}[c]{@{}l@{}}accaduto \\ \textit{\small{[}'happened'{]}}\end{tabular} & \cellcolor[HTML]{E67C73}\small{-0.17} & \begin{tabular}[c]{@{}l@{}}della \\ \textit{\small{[}'of the' (+ f.noun){]}}\end{tabular} & \cellcolor[HTML]{E8928A}\small{-0.08} & \begin{tabular}[c]{@{}l@{}}trovata \\ \textit{\small{[}'found' (ptc, f.sg.){]}}\end{tabular} & \cellcolor[HTML]{F9E3E1}\small{-0.20} &  & \begin{tabular}[c]{@{}l@{}}ospedale\\ \textit{\small{[}'hospital'{]}}\end{tabular} & \cellcolor[HTML]{E67C73}\small{-0.16} & cx:Buildings++nvrb & \cellcolor[HTML]{E99A93}\small{-0.07} & \begin{tabular}[c]{@{}l@{}}auto\\ \textit{\small{[}'car'{]}}\end{tabular} & \cellcolor[HTML]{F5CDC9}\small{-0.41} \\
-1 & . & \cellcolor[HTML]{E67C73}\small{-0.35} & . & \cellcolor[HTML]{E67C73}\small{-0.13} & \begin{tabular}[c]{@{}l@{}}morta \\ \textit{\small{[}'dead' (f.sg.){]}}\end{tabular} & \cellcolor[HTML]{F9E1DF}\small{-0.21} &  & \begin{tabular}[c]{@{}l@{}}due\\ \textit{\small{[}'two'{]}}\end{tabular} & \cellcolor[HTML]{E67C73}\small{-0.16} & frm:Buildings & \cellcolor[HTML]{E7857D}\small{-0.09} & \begin{tabular}[c]{@{}l@{}}uccide \\ \textit{\small{[}'he/she/it kills'{]}}\end{tabular} & \cellcolor[HTML]{F5CBC8}\small{-0.42} \\ \bottomrule
\end{tabular}%
}\caption{Comparison of most informative features for an `easy' attribute (blame/murderer) and a `hard' attribute (focus/concept). \textit{\textit{\small{[}Abbreviations: rol=semantic role, frm=frame, cx=construction, nvrb=nonverbal, Pers\_rel=Personal\_relationship; f.=feminine [grammar], ptc.=participle, sg.=singular, pl.=plural, rel.pn.=relative pronoun, cmp.=complementizer{]}}}}\label{tab:explanation}
\end{table*}

This pattern is also reflected in MSA 
(Table~\ref{tab:maxcol}): for \textit{Focus}, it is substantially easier to predict the dimension with the highest score than for \textit{Blame} and \textit{Cause}. However, all models perform well above chance level for each of the questions, with the strongest overall scores for BERTino (56-72\%).\\
\indent The gain in performance achieved by the BERT-based models with respect to the surface feature models varies substantially between attributes. For example, the \textit{bow} model has a surprisingly high score for \textit{blame on murderer} (R$^2$ 0.49), with only moderate gains from the BERT-IT and BERTino models (resp. +0.06 and +0.12 points). By contrast, \textit{bow} scores poorly on \textit{focus on concept} (R$^2$ 0.13), whereas BERT-IT and BERTino have good scores (R$^2$ 0.63/0.64). To get additional insight into the differences between models, we performed a feature attribution analysis. For the \textit{bow} and \textit{f1+} ridge regression models, we simply extracted the feature weights with the lowest and highest absolute values; for transformers, we applied the \textit{integrated gradients} interpretation method \cite{SundararajanTY17}\footnote{We used the implementation provided by the \textit{transformers-interpret} package, see \url{https://github.com/cdpierse/transformers-interpret}} to obtain token-based attribution values for all sentences in the test set, and used the averaged values for tokens above a frequency threshold ($k\geq 5$, on a test set of 300 sentences) as an approximation of the overall feature importance. The results for \textit{blame on murderer} and \textit{focus on concept} are shown in Table~\ref{tab:explanation}. For \textit{blame on murderer}, all three models seem to focus on similar lexical items: for example, ``\textit{uccide}'' (`(he) kills') has a high positive attribution value in both the \textit{bow} ridge regression and the fine-tuned BERTino model, and in \textit{f1+} we find a positive score for the \fnframe{Killing} frame, which is an abstraction over killing-related words. We also find that personal relationships (`wife', `ex', \fnframe{Personal\_relationship}) get positive attributions in all three models.
By contrast, we find negative attribution values for ``\textit{accaduto}'' (`happened') and the corresponding \fnframe{Event} frame in \textit{bow} and \textit{f1}, which maps neatly onto our observations discussed in \S\ref{sub:questionnaire-results}.  
For \textit{focus on concept}, no insightful differences between the three models are immediately obvious. We do find several intuitively relevant features in each model: ``passionale'' (`out of passion') and  ``femminicidio'' (`femicide') could to examples of concepts that sentences could give focus to, whereas ``omicida'' (`murderer/murderous') and ``killer'' could be seen as emphasizing the role of a human agent rather than an abstract concept. 
\section{Conclusion \& Future Work}
\label{sec:conclusion}

This paper has presented a detailed analysis of human perceptions \rebuttal{of responsibility in Italian news reporting on GBV}. The judgments we collected confirm the findings of previous work on the impact of specific grammatical constructions and semantic frames, and the perceptions they trigger in readers. 

On the basis of the results of our survey, we have investigated to what extent different NLP architectures can predict the human perception judgements. The results of our experiments indicate that fine-tuning monolingual transformers leads to the best results across multiple evaluation measures. This opens up the possibility of integrating systems able to identify potential perception effects as support tools for media professionals. 

In the future, we plan to run a more detailed analysis of the data considering differences along individual and demographic dimensions of the respondents. In addition to this, natural follow-up experiments will focus on the application of the approach to other languages and cultural contexts both targeting GBV as well as other socially relevant topics, e.g. car crashes~\cite{te2020framing}.

%
\section*{Ethics Statement}

\paragraph{Limitations}

This work has a strong connection with multiple theoretical frameworks: Frame Semantics, Construction Grammar, and Critical Discourse Analysis. The way we have structured the questionnaire aimed at collecting data from human participants with respect to different sentences - which in different ways contained variations in syntactic structures and semantic frames that could be linked to findings and claims about the ``\textit{perception}'' and its effects in the interpretation of sentences. The use of state-of-the-art NLP tools to identify these properties in a large collection of data represents both an advantage (i.e., allows to deal with a large number of data, reducing human subjective interpretation) and a limit (i.e., errors from the systems may result in non optimal examples for human judgements).

While representing an \textit{unicum} in the language resource panorama, since there are no previous comparable and available corpora, the number of available sentences used to train the models is somewhat limited. The final corpus, however, represents an optimal compromise between number of judgements needed to obtain a solid representations of perceptions by users and number of data points that could be used by stochastic NLP architectures to learn from the data. 

Finally, the outcome of the perception judgements can be generalized to the population of Italian young adults attending universities (i.e., undergrad students). This is a limitation of the data collection process. We tried to minimize this by reaching out to students in multiple universities (i.e., geographical variation) and at different faculties (from Arts/Humanities, to Computer Science and Physics) and disciplines (from Linguistics, to Media and Communication Studies, Computer Science, and Physics).  

\paragraph{Data collection}
The questionnaire was conducted using the Qualtrics XM platform. Participation to the questionnaire was on a voluntarily basis. Participants could interrupt their participation in any moment. Only fully completed questionnaires have been retained. Participants received compensation (5 euros) - upon completion of the questionnaire. Participants have been recruited mainly among undergraduate students at different universities in Italy. 

Participation was fully anonymous: 1) participants could access the questionnaire via a unique special access token that could be obtained by filling in a form; 2) no personal information other than the participants' email address was stored; 3) IP addresses were not stored or tracked; 4) the special access token and the participants' email were decoupled. Participants could receive their compensation only by providing the unique access token. 

\paragraph{Dual use}
The experiments we have run investigate to what extent models are able to predict human perceptions along three dimensions with respect to GBV. The very nature of the task limits the potential misuse by malevolent agents. At the same time, malevolent agents can purposefully misrepresent the results to minimize the negative aspects associated to the reporting of the phenomenon by media. By making the models and the data publicly available, together with a detailed explanation of how the models work and how results should be interpreted in a correct way, we mitigate these risks.

\paragraph{Intended use}
As it is the case for supervised models, sensitivity to the training material is high. At the moment, we have not tested the portability of the models to other topics. We do recommend to use these models only on data compatible with the phenomenon we have taken into account, i.e., GBV against women. Although the application of the models to any other type of texts reporting violence and killing against other targets may still give some valid results, we discourage its use since risks of unforeseen behaviors are high, with potential harmful consequences for the victims of violence.

\section*{Acknowledgements}
The research reported in this article was funded by the Dutch National Science organisation (NWO) through the project \textit{Framing situations in the Dutch language}, VC.GW17.083/6215. We thank the Center for Information Technology of the University of Groningen for their support and for providing access to the Peregrine high performance computing cluster.

\bibliography{bib/anthology,bib/custom}
\bibliographystyle{acl_natbib}

\end{document}